\documentclass[conference]{IEEEtran}
\IEEEoverridecommandlockouts
\usepackage{cite}
\usepackage{amsmath,amssymb,amsfonts}
\usepackage{graphicx}
\usepackage{textcomp}
\usepackage{xcolor}
\usepackage[american]{babel}
\usepackage[T1]{fontenc}
\usepackage{fancyhdr}
\usepackage{tikz}

\usepackage{algorithm}
\usepackage{algpseudocode}
\usepackage{amsfonts}
\usepackage{multirow}
\usepackage{lipsum}
\usepackage{balance}
\usepackage{subcaption}
\usepackage{array}
\usepackage{svg}
\usepackage{adjustbox} 
\usepackage[table,xcdraw]{xcolor}
\usepackage{pifont}
\newcommand{\cmark}{\ding{51}} 
\newcommand{\xmark}{\ding{55}} 
\usepackage{booktabs}

\def\BibTeX{{\rm B\kern-.05em{\sc i\kern-.025em b}\kern-.08em
    T\kern-.1667em\lower.7ex\hbox{E}\kern-.125emX}}
    
\begin{document}


\title{CAMP-HiVe: Cyclic Pair Merging based Efficient DNN Pruning with Hessian-Vector Approximation for Resource-Constrained Systems \vspace{-3mm}}

\author{\IEEEauthorblockN{
\large Mohammad Helal Uddin\IEEEauthorrefmark{1}, Sai Krishna Ghanta\IEEEauthorrefmark{2}, Liam Seymour\IEEEauthorrefmark{1},  Sabur Baidya\IEEEauthorrefmark{1}\\
\IEEEauthorblockA{\IEEEauthorrefmark{1}\normalsize Department of Computer Science and Engineering, University of Louisville, KY, USA}
\IEEEauthorblockA{\IEEEauthorrefmark{2}\normalsize School of Computing, University of
Georgia, Athens, GA, USA}
\normalsize {e-mail: mohammad.helaluddin@louisville.edu, sai.krishna@uga.edu, wrseym03@louisville.edu, sabur.baidya@louisville.edu}
\vspace{-4mm}
}}

\maketitle

\pagestyle{fancy}
\thispagestyle{fancy}
\renewcommand{\headrulewidth}{0pt}  
 
\fancyhf{}
 
\fancyhead[C]{
    \small This article has been accepted for publication in the IEEE International Conference on Machine Learning and Applications (ICMLA) 2025. 
}
 
\fancyfoot[C]{
    \begin{tikzpicture}[remember picture, overlay]
        \node[anchor=south,yshift=10pt] at (current page.south) {
            \parbox{\textwidth}{
                \centering
                \footnotesize
                \textcopyright~2025 IEEE. Personal use of this material is permitted. Permission from IEEE must be obtained for all other uses, in any current or future media, including reprinting/republishing this material for advertising or promotional purposes, creating new collective works, for resale or redistribution to servers or lists, or reuse of any copyrighted component of this work in other works.
            }
        };
    \end{tikzpicture}
}

\begin{abstract}
Deep learning algorithms are becoming an essential component of many artificial intelligence (AI) driven applications, many of which run on resource-constrained and energy-constrained systems. For efficient deployment of these algorithms, although different techniques for the compression of neural networks models are proposed, neural pruning is one of the fast and effective methods which can provide a high compression gain with minimal cost. To harness  enhanced performance gain with respect to model complexity, we propose a novel neural network pruning approach utilizing Hessian-vector products that approximates crucial curvature information in the loss function which significantly reduces the computation demands. By employing a power iteration method, our algorithm effectively identifies and preserves the essential information, ensuring a balanced trade-off between model accuracy and computational efficiency. 
Herein, we introduce CAMP-HiVe, a cyclic pair merging based pruning with Hessian Vector approximation by iteratively consolidating weight pairs, combining significant and less significant weights, thus effectively streamlining the model while preserving its performance. This dynamic, adaptive framework allows for real-time adjustment of weight significance, ensuring that only the most critical parameters are retained. Our experimental results demonstrate that our proposed method achieves significant reductions in computational requirements while maintaining high performance across different neural network architectures, e.g., ResNet18, ResNet56, and MobileNetv2 on standard benchmark datasets, e.g., CIFAR10, CIFAR-100, and ImageNet, and it outperforms the existing state-of-the-art neural pruning methods.
\end{abstract}


\vspace{-2mm}
\section{\textbf{Introduction}}
\vspace{-1mm}
Recent advancements in deep learning \cite{b1} \cite{b2},
have significantly boosted the accuracy of tasks like computer vision \cite{b3}\cite{b4} and natural language processing \cite{b5}\cite{b6}. A key driver behind these advancements is the scale of the models employed, often boasting millions or even billions of trainable parameters \cite{b7} enabling complex model architecture. The scale of these models 
is expected to further increase
in the foreseeable future.
But there are major performance issues with deploying such big models in the real world.
Hence, scientists have been looking for ways to make these deep neural networks that rely on a lot of parameters, smaller and lighter so that they are easier to deploy. Considering the adaptations from the Optimal Brain Damage/Surgeon (OBD/OBS) framework \cite{b8}, the common solution is to boost deep neural networks by removing the less significant connections, which have a minute impact on the network’s performance. However, one important hurdle persists: the lack of a global compression approach that allows for the effective re-using, and adaptation of existing models without extensive training \cite{b9}. Bridging this gap is crucial for fully unlocking the potential of pre-trained models and streamlining their broad utilization across various fields and objectives.

In this study, we propose a 
new 
pruning technique with weight redistribution and merging for compressing neural networks . This approach provides a versatile balance between efficiency and performance quality.
The primary requirement of this approach is the precise estimation of the hessian vector products. The Hessian vector product  reduces the computation of the Hessian's effect on a vector without the need to fully calculate or store the entire matrix. Moreover, The Hessian product vectors are a core component in several parts of deep learning \& optimization \cite{b10}. The application of second-order approximations for large-scale models can be challenging, leading to the frequent use of coarse-grained approximations such as Kronecker-factorization\cite{b10}. Though there is a limited knowledge regarding the effectiveness and scalability of these approaches.

\begin{figure}[!t]
    \centering
    \vspace{-3mm}
    \includegraphics[scale=0.27]{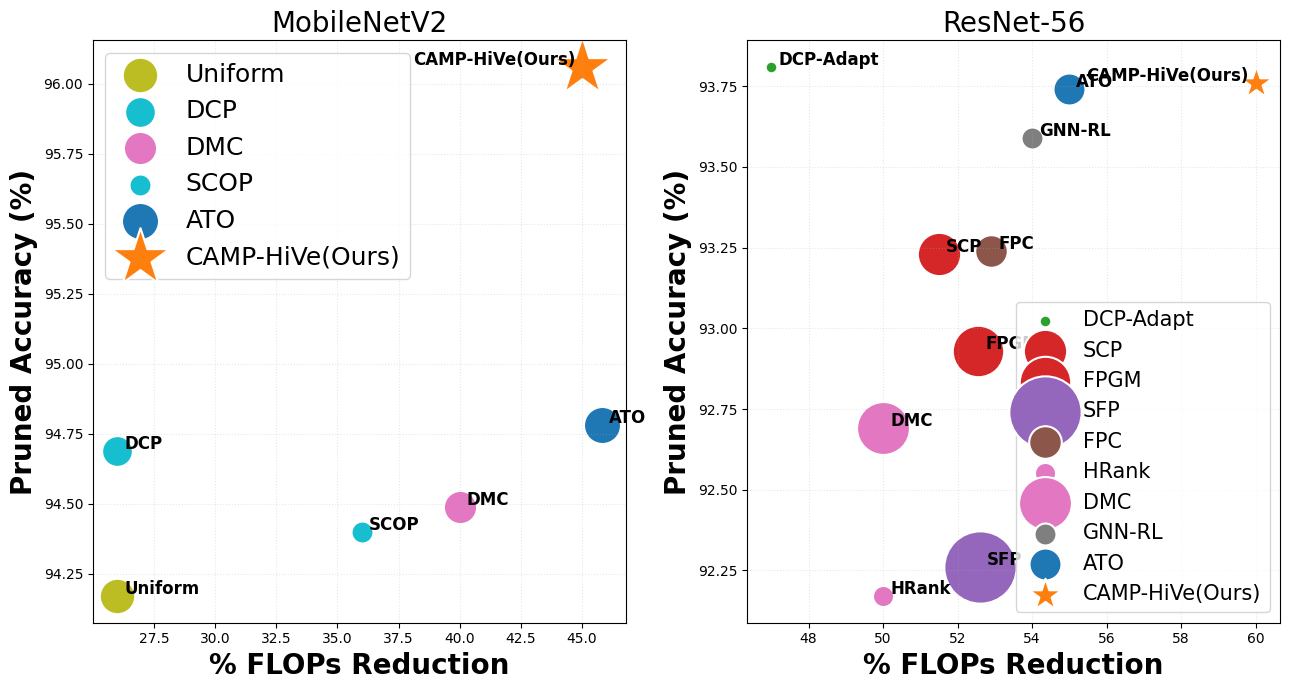}
    \vspace{-7mm}
    \caption{\small{Accuracy vs. FLOPs Reduction Comparision between different models (Bubble $\propto |\Delta\mathrm{-Acc}|$). More detailed results have been presented in Tables \ref{Cifat10_100_DataResult} and \ref{imagenetDataResult}.}}
\label{fig:parato_v2_ref}
\vspace{-7mm}
\end{figure}

\vspace{1mm}
\noindent
\textbf{Motivation:} The redundancy in the parameterization of deep models can lead to unnecessary computational expense and memory usage. This research aims to address these challenges by proposing an innovative weight pruning methodology that leverages the Hessian-vector product to estimate the significance of weights within a neural network.


\begin{table*}[!ht]
\centering
\resizebox{\textwidth}{!}{%
\begin{tabular}{lccccccc}
\toprule
\textbf{Year} & \textbf{2017} & \textbf{2018} & \textbf{2019} & \textbf{2020} & \textbf{2021} & \textbf{2023} & \textbf{2024/25} \\
\textbf{Method} & Molchanov17\cite{b49} & Uniform\cite{b30} & AMC\cite{b45} & HRank\cite{b17} & ATO\cite{b25} & EigenDamage\cite{b20} & CAMP-HiVe(Ours) \\
\midrule
\multicolumn{8}{l}{\textbf{Core Techniques}} \\
Hessian‐based Sensitivity & \xmark & \xmark & \xmark & \xmark & \cmark & \cmark & \cmark \\
Second‐order Approximation & \cmark & \xmark & \xmark & \cmark & \cmark & \cmark & \cmark \\
\midrule
\multicolumn{8}{l}{\textbf{Objective / Loss‐level Innovations}} \\
Taylor‐series $\Delta\mathcal{L}$ minimisation & \cmark & \xmark & \xmark & \xmark & \xmark & \cmark & \cmark \\
Weight‐merging loss preservation & \xmark & \xmark & \xmark & \xmark & \xmark & \xmark & \cmark \\
Multi‐objective (Acc+FLOPs+Latency/Power) & \xmark & \xmark & \cmark & \xmark & \cmark$^{\dagger}$ & \xmark & \cmark \\
\midrule
\multicolumn{8}{l}{\textbf{Empirical Evaluation Paradigms}} \\
Inference latency analysis & \xmark & \xmark & \xmark & \xmark & \xmark & \xmark & \cmark \\
Edge–device deployment & \xmark & \xmark & \xmark & \xmark & \xmark & \xmark & \cmark \\
Power consumption analysis & \xmark & \xmark & \xmark & \xmark & \xmark & \xmark & \cmark \\
Detailed feature‐map analysis & \xmark & \xmark & \xmark & \xmark & \xmark & \xmark & \cmark \\
\bottomrule
\end{tabular}%
}
\vspace{-2mm}
\caption{\small{Comparative novelty timeline highlighting objective‐level contributions of our Cyclic Pair Merge Pruning (CAMP-HiVe) relative to representative pruning methods. $^\dagger$ ATO optimise a Controller reward that balances accuracy and FLOPs, but do not include measured latency or power in the evaluation.}}
\label{tab:comparative_novelty_timeline}
\vspace{-7mm}
\end{table*}




\vspace{1mm}
\noindent
\textbf{Contribution:}
We introduce a novel pruning technique that identifies less significant weights via Hessian-vector product calculations, a method that approximates the impact of each weight on the learning outcome. Unlike traditional pruning methods that may indiscriminately remove weights, our approach merges these less significant weights with more critical ones before pruning, ensuring that the overall network performance is preserved or even enhanced. This method not only reduces the model's complexity significantly by decreasing the number of floating-point operations (FLOPs) but also retains, if not improves, the accuracy of the network. CAMP-HiVe outperforms SOTA baselines (Figure~\ref{fig:parato_v2_ref}), and Table~\ref{tab:comparative_novelty_timeline} highlights our novel contributions in the context of recent advancements.

\section{\textbf{Related Works}}
\vspace{-1mm}

The Lottery Ticket Hypothesis \cite{b11} has introduced the idea that inside a bigger neural network, there is smaller subnetwork which if trained independently, can perform just as well as the whole thing. \cite{b12} explored whether this theory holds true when scaled up to bigger models and larger datasets, while \cite{b13} developed Iterative Magnitude Pruning (IMP) with weight rewinding, a clever technique to more efficiently pinpoint these subnetworks.However, these approaches typically involve extensive retraining, which can become computationally expensive.

To achieve hardware-friendly sparsity, structured pruning techniques have continued to advance by carefully removing entire filters, channels, or even layers. One of the notable work was Soft Pruning(SFP)\cite{b16} which involves pruning filters during training and allowing them to recover.  Similarly, the HRank\cite{b17} evaluates filters based on the intrinsic structure of their feature maps, achieving noteworthy efficiency improvements with modest computational demands. 
These techniques increase model efficiency, but if they are not used carefully, they may still struggle with considerable accuracy drops.

Pruning has been coupled with knowledge distillation to enhance the performance of pruned models. DistilBERT, a condensed and distilled version of BERT from the original, larger model, was proposed by \cite{b18}. In a more recent work, \cite{b19} presented TinyBERT, which transfers knowledge from a large teacher model to a smaller student model while removing weights that are not as significant through a two-stage distillation process. Although this method adds complexity to the training pipeline, it successfully balances performance retention and model size reduction.

There have been notable developments in second-order \& third-order pruning techniques, using the Hessian matrix for sensitivity analysis. \cite{b20} presented EigenDamage, a pruning technique that achieves state-of-the-art results with little degradation in performance by using the dominant eigenvalues of the Hessian to guide pruning decisions. \cite{b21}, introduced a scalable technique for approximating the Hessian, which allows for second-order pruning on networks of high scale. Although these techniques offer accurate pruning criteria, they are computationally demanding and require effective approximations, which we address and overcome in our approach.

\section{\textbf{Technical Approach}}
\vspace{-1mm}
\subsection{The Problem Settings and Motivation}
\vspace{-1mm}
We consider a supervised learning, where we are given a dataset $\mathcal{D}$ composed of $n$ datapoints $(\mathbf{x}_i, a_i, y_i)$, where each datapoint is independently and identically distributed from an unknown distribution $\Pi$. Each datapoint consists of a feature vector $\mathbf{x}_i \in \mathcal{X}$, and a class label $y_i \in \mathcal{Y}$. The empirical loss function, calculated as follows:
\vspace{-3mm}
\begin{equation}
\mathcal{L}(\mathbf{w}) = \frac{1}{n} \sum_{i=1}^{n} \mathcal{L}_i(\mathbf{x}_i, y_i, \mathbf{w}),
\vspace{-4mm}
\end{equation}

The proposed pruning scheme adjusts the weights within the architecture for better trade-off between the accuracy and latency via preserving essential information. The performance variation can be determined by the change in the loss function $\mathcal{L}$. The primary objective is to minimize the variation in the loss function as follows:
\begin{equation}
\begin{split}
\min_{\Delta \mathbf{w}} \Delta \mathcal{L}(\mathbf{w}, \Delta \mathbf{w}) &= \mathcal{L}(\mathbf{w} + \Delta \mathbf{w}) - \mathcal{L}(\mathbf{w}), \\
\textit{subject to } & \mathcal{C}(\mathbf{w} + \Delta \mathbf{w}) \leq \mathcal{C}_{\text{budget}}
\end{split}
\end{equation}

\subsection{\textbf{Hessian Matrix Approximation}}
\vspace{-2mm}
The proposed merge pruning focuses on the relative importance of model weights and adjusts them to achieve a better trade-off between accuracy and latency via preserving essential information. In order to calculate the loss,  $\Delta \mathcal{L}$, we have used a modified version of third-order Taylor expansion around $\mathbf{w}$ as follows,
\vspace{-2mm}
\begin{equation}
\begin{aligned}
\Delta \mathcal{L} & \approx \Delta \mathbf{w}^T \mathbf{g} + \frac{1}{2} \Delta \mathbf{w}^T \mathbf{Hv} + \\
& \frac{1}{6} \sum_{i,j,k} \frac{\partial^3 \mathcal{L}}{\partial w_i \partial w_j \partial w_k} \Delta w_i \Delta w_j \Delta w_k
\end{aligned}
\vspace{-2mm}
\end{equation}

Here $\mathbf{g}$ and $\mathbf{H}\mathbf{v}$ represent the gradient of the loss function and the Hessian vector product at $\mathbf{w}$. This product provides the multiplication of the Hessian matrix \(H\) at \(\mathbf{w}\) with a vector \(\mathbf{v}\), which can be taken as the update step \(\Delta \mathbf{w}\), without explicitly forming the full Hessian matrix. According to quadratic approximation, the third term in Equation 3 can be neglected. 
For large models, directly computing and storing this matrix is computationally prohibitive due to its size ($n \times n$) and the need to calculate every second-order partial derivative.



The Hessian-vector product $Hv$ referred in \cite{b22}\cite{b23}\cite{b24}, where $v$ is a vector, is defined as:

\vspace{-3mm}
\[
\hspace{6mm}
Hv = H \cdot v = \begin{bmatrix} \sum_{j=1}^{n} H_{1j} v_j \\ \sum_{j=1}^{n} H_{2j} v_j \\ \vdots \\ \sum_{j=1}^{n} H_{nj} v_j \end{bmatrix}
\vspace{-1mm}
\]

This product can be computed efficiently without explicitly forming the entire Hessian matrix. This efficiency arises as the Hessian-vector product can be computed using automatic differentiation techniques, which are similar in complexity to computing a gradient. Hessian vector product reduces the calculation of all the second-order partial derivatives.

 Hessian matrix is computationally heavy, to avoid this, we have used Hessian vector products $Hv$. It can get the important information of curvature without needing the whole Hessian.To achieve this, we employ a numerical approximation using a small scalar $\epsilon$ and perform power iteration on initialized vectors associated with the weights.The power iteration method is employed to approximate the Hessian-vector by utilizing initialized vectors matching to the weights. 
\vspace{-1mm}
\begin{equation}    \mathbf{v}_{\text{new}}^{(k)} = \frac{1}{\epsilon} \left( \nabla \mathcal{L}(\mathbf{w}^{(k)} + \epsilon \mathbf{v}^{(k)}) - \nabla \mathcal{L}(\mathbf{w}^{(k)}) \right)
\vspace{-2mm}
\end{equation}
\begin{equation}    \mathbf{v}^{(k)}_{\text{new}} \leftarrow \text{normalize}\left(\frac{1}{\epsilon} \left( \nabla \mathcal{L}(\mathbf{w}^{(k)} + \epsilon \mathbf{v}^{(k)}) - \nabla \mathcal{L}(\mathbf{w}^{(k)}) \right)\right)
\vspace{-2mm}
\end{equation}

The power iteration process repeatedly applies the Hessian-vector product to approximate the dominant eigenvector, which helps in capturing the most significant curvature details of the loss landscape. The significance of $j$th weight $\sigma_j^{(k)}$ in $k$-th layer approximated as the magnitude of the corresponding components in the approximated eigenvectors.  i.e, $\sigma_j^{(k)} = \left| v_{\text{new},j}^{(k)} \right|$. 
The threshold $\theta$ for pruning and merging the weights is estimated based on the significance of weights and a pre-determined percentile $p$. 
\begin{equation}
    \theta = Q_p\left( \left\{ \left| v_{\text{new},j}^{(k)} \right| : j = 1, \dots, n_k \right\} \right),
    \vspace{-2mm}
\end{equation}

where $Q_p(\cdot)$ denotes the $p$-th percentile function, and $n_k$ is the number of weights in layer $k$. The weights are partitioned into  $S$ ($\sigma_j^{(k)}) \geq \theta$) 
and $L$ ($\sigma_j^{(k)}) < \theta$) based on their significance. Now merging happens between the less significant and significant weights using the cyclic pair technique.
\begin{equation}
    \forall i \in L, \quad w^{(k)}_{\text{sig}} += w^{(k)}_i, \quad w^{(k)}_i = 0.
    \vspace{-3mm}
\end{equation}

Finally, the binary mask was utilized in order to bring the weights up to date. This was done in order to guarantee that only the weights that were significant and accurately redistributed remained under consideration.
\vspace{-2mm}
\begin{equation}    \mathbf{W}_{\text{pruned}}^{(k)} = \mathbf{W}^{(k)} \odot \mathbf{m}^{(k)} 
    \vspace{-1mm}
\end{equation}


\subsection{\textbf{Proposed Solution: Cyclic Pairing and Merging}}

\vspace{-1mm}

In the process of cyclic pairing, every individual less significant weight \( W_{\text{less},i} \) get paired with a significant weight using a cycli index:
\vspace{-2mm}
\begin{equation}
    j = (i \mod s) + 1 
    \vspace{-3mm}
\end{equation}

This pairing technique ensures that every significant weight receives contributions from numerous less significant weights, so preserving a balance even when \( l \) is not a multiple of \( s \). For each paired weight \( (W_{\text{sig},j}, W_{\text{less},i}) \):
\vspace{-2mm}
    \begin{equation}
    W'_{\text{sig},j} = W_{\text{sig},j} + W_{\text{less},i} 
    \label{eqn:wt_update}
    \vspace{-2mm}
    \end{equation}
    \begin{equation}
        W'_{\text{less},i} = 0 
        \label{eqn:wt_zer0}
        \vspace{-1mm}
    \end{equation}
    Equation~\ref{eqn:wt_update} updates the significant weight by adding the value to its paired less significant weight. Equation~\ref{eqn:wt_zer0} updates the less significant weight as zero after its contribution, signifying that its value has been fully transferred.

\noindent
{\bf Complete Update Form}

The entire update process across all weights can be succinctly represented as:
\[
\vspace{-2mm}
\begin{split}
W'_{\text{sig},(i \mod s) + 1} = W_{\text{sig},(i \mod s) + 1} + W_{\text{less},i} \quad \\
\text{for } i = 1, 2, \ldots, l 
\end{split}
\]
\[
\vspace{-1mm}
W'_{\text{less},i} = 0 \quad \text{for } i = 1, 2, \ldots, l 
\]

\subsection{\textbf{Information Preservation}}
\vspace{-2mm}
In neural networks, the standard practice of magnitude-based pruning directly zeros out weights in the weight matrix $W$ that are below a predefined threshold $\epsilon$:
\vspace{-2mm}
\begin{equation}
    \text{if } |W_{ij}| < \epsilon, \text{ then } W'_{ij} = 0; \text{ otherwise } W'_{ij} = W_{ij}.
    \vspace{-3mm}
\end{equation}

This can result in significant information loss, particularly when small trimmed weights are crucial to the network's learned features.
Our cyclic pair merging pruning method uses the dominant eigenvector of Hessian vector products to account for the second-order sensitivity of the loss function to weight changes.
\vspace{-4mm}
\begin{equation}
    H_{ij} = \frac{\partial^2 L}{\partial W_i \partial W_j}.
    \vspace{-3mm}
\end{equation}

With the help of $v$ we can pinpoint the weights that greatly influence the loss function. These weights are subsequently arranged based on the magnitude of their entries in $v$ and merged cyclically. Specifically, less significant weights $W_{ij}$ are merged to more significant weights $W_{kl}$:
\vspace{-2mm}
\begin{equation}
W'_{kl} = W_{kl} + W_{ij}, \quad W'_{ij} = 0.
\vspace{-3mm}
\end{equation}

This process redistributes the values of less significant weights into significant ones, thus preserving the total information in the layer. It ensures that important characteristics of the loss landscape, captured by the curvature indicated by $v$, are retained. 

\setlength{\textfloatsep}{0pt}
\begin{algorithm}[!t]
\footnotesize
\caption{Cyclic Pair Merging Pruning}
\begin{algorithmic}[1]
\Require Pre-trained model $N$, dataset $D$ , threshold \(\epsilon\)
\Ensure A pruned and adjusted neural network $N'$.

\State \textbf{Initialize:}
\State \text{Load and preprocess dataset}
\State \text{Load pre-trained model}

\State \textbf{Compute FLOPS:}
\For{each $l$ in model}

    \State \hspace{-4mm} $\text{FLOPS}[l] \gets \begin{cases} 
        2 \times o_h \times o_w \times o_c \times (k_h \times k_w \times i_c) & \\\phantom{aaaaaaaaaaaa} \text{if } l \text{ is Conv2D} \\ 
        2 \times i_s \times o_s \phantom{aaa}
        \text{if } l \text{ is Dense} 
    \end{cases}$
\EndFor

\State \textbf{Power Iteration for Eigenvector:}
\For{$t = 1$ to $\text{max\_iterations}$}
    \State $v \gets \text{random\_vector}()$
    \For{each $(x\_batch, y\_batch)$ in dataset}
        \State $\nabla \gets \text{compute\_gradients}(\text{$N$}, x\_batch, y\_batch)$
        \State $v \gets \text{normalize}(\text{compt\_hessian\_vect\_prod}(\nabla, v))$
    \EndFor
\EndFor

\State \textbf{Weight Selection and Sorting:}
\For{each $l$ in $N$}
    \State $W \gets l.\text{get\_weights}()$
    \State \text{sort } $W$ \text{ based on } $v$
    \State $(W_{\text{sig}}, W_{\text{less}}) \gets \text{partition}(W, \epsilon)$
\EndFor

\State \textbf{Cyclic Pairing:}
\State $\{(W_{\text{less},i}, W_{\text{sig},(i \mod |W_{\text{sig}}|) + 1}) \}_{i=1}^{|W_{\text{less}}|}$

\State \textbf{Merge and Prune:}
\For{each $(W_{\text{sig},j}, W_{\text{less},i})$}
    \State $W'_{\text{sig},j} \gets W_{\text{sig},j} + W_{\text{less},i}$
    \State $W'_{\text{less},i} \gets 0$
    \State \text{Update model weights}
\EndFor

\State \textbf{Fine-tuning the Pruned and Adjusted Network}
\State \textbf{if} fine-tuning is desired \textbf{then}
\State \hspace{1.2cm} Retrain $N'$ using dataset $D$
\State \hspace{1.2cm} \textbf{while} not converged \textbf{do}
\State \hspace{2.2cm} Update weights in $N'$.
\State \hspace{1.2cm} \textbf{end while}
\State \textbf{end if}
\State \textbf{return} $N'$

\end{algorithmic}
\end{algorithm}

\vspace{1mm}
\section{\textbf{Experiments and Results}}\label{sec:math}
\vspace{-2mm}
Here we first discuss the datasets and experimental details. 
Then we compare the efficacy of our proposed Cyclic-paired merge pruning with state-of-the-art approaches. 

\vspace{1mm}
\textbf{Baseline Training:}
 We used several datasets, e.g., CIFAR10, CIFAR100, Imagenet in our experiments.We adopt popular neural network architectures, e.g., MobilenetV2, Resnet56, Resnet-18. For CIFAR-10 and CIFAR-100 datasets, we follow standard training for 100 epochs and a learning rate of 0.001.



\begin{center}
\begin{table*}[!t]
\centering
\renewcommand{\arraystretch}{1} 
\tiny
\resizebox{\textwidth}{!}{%
\begin{tabular}{|c|c|c|c|c|c|c|}
\hline
Dataset & Architecture & Method & Baseline Accuracy & Pruned Accuracy & $\Delta$-Acc & $\downarrow$ FLOPs \\ \hline
\multirow{14}{*}{CIFAR-10} 
& \multirow{2}{*}{ResNet-18} & OTOv2\cite{b28} & 93.02\% & 92.86\% & -0.16\% & 79.7\% \\ \cline{3-7}
&  & ATO\cite{b25} & 94.41\% & 94.51\% & +0.10\% & 79.8\% \\ \cline{3-7}
&  & \textbf{CAMP-HiVe(Ours)} & \textbf{96.04\%} & \textbf{96.47\%} & \textbf{+0.43\%} & \textbf{80\%} \\ \cline{2-7}

& \multirow{10}{*}{ResNet-56} 
& DCP-Adapt\cite{b30} & 93.80\% & 93.81\% & +0.01\% & 47.0\% \\ \cline{3-7}
&  & SCP\cite{b31}& 93.69\% & 93.23\% & -0.46\% & 51.5\% \\ \cline{3-7}
&  & FPGM\cite{b32} & 93.59\% & 92.93\% & -0.66\% & 52.56\% \\ \cline{3-7}
&  & SFP\cite{b33} & 93.59\% & 92.26\% & -1.33\% & 52.6\% \\ \cline{3-7}
&  & FPC\cite{b34} & 93.59\% & 93.24\% & -0.25\% & 52.9\% \\ \cline{3-7}
&  & HRank\cite{b17} & 93.26\% & 92.17\% & -0.09\% & 50.0\% \\ \cline{3-7}
&  & DMC\cite{b36} & 93.62\% & 92.69\% & +0.7\% & 50.0\% \\ \cline{3-7}
&  & GNN-RL\cite{b37}  & 93.49\% & 93.59\% & +0.10\% & 54.0\% \\ \cline{3-7}
&  & ATO\cite{b25} & 93.50\% & 93.74\% & +0.24\% & 55.0\% \\ \cline{3-7}
&  & \textbf{CAMP-HiVe(Ours)} & \textbf{93.49\%} & \textbf{93.76\%} & \textbf{+0.27\%} & \textbf{60\%} \\ \cline{2-7}
& \multirow{6}{*}{MobileNetV2} 
& Uniform \cite{b30}& 94.47\% & 94.17\% & -0.30\% & 26.0\% \\ \cline{3-7}
&  & DCP \cite{b30}& 94.47\% & 94.69\% & +0.22\% & 26.0\% \\ \cline{3-7}
&  & DMC \cite{b36} & 94.23\% & 94.49\% & +0.26\% & 40.0\% \\ \cline{3-7}
&  & SCOP\cite{b40} & 94.5\% & 94.4\% & -0.1\% & 36.0\% \\ \cline{3-7}
&  & ATO\cite{b25} & 94.45\% & 94.78\% & +0.33\% & 45.8\% \\ \cline{3-7}
&  & \textbf{CAMP-HiVe(Ours)} & \textbf{95.1\%} & \textbf{96.06\%} & \textbf{+0.93\%} & \textbf{45\%} \\ \hline
\multirow{2}{*}{CIFAR-100} 
& \multirow{2}{*}{ResNet-18} & OTOv2\cite{b28} & - & 74.96\% & - & 39.8\% \\ \cline{3-7}
&  & ATO\cite{b25}& 77.95\% & 76.79\% & -0.07\% & 40.1\% \\ \cline{3-7}
&  &\textbf{CAMP-HiVe(Ours)} & \textbf{80.80\%} & \textbf{83.59\%} & \textbf{+2.79\%} & \textbf{50\%} \\ \hline
\end{tabular}%
}
\vspace{-1mm}
\caption{Detailed accuracy and FLOPs reduction for different architectures on CIFAR-10 and CIFAR-100 datasets.}
\label{Cifat10_100_DataResult}
\vspace{-2mm}
\end{table*}
\end{center}
\vspace{-8mm}

\begin{table*}[!t]
\centering
\renewcommand{\arraystretch}{1.5} 
\resizebox{\textwidth}{!}{%
\begin{tabular}{|c|c|c|c|c|c|c|}
    \hline
    \textbf{Architecture} & \textbf{Method} & \textbf{Base Top-1} & \textbf{Base Top-5} & \textbf{Pruned Top-1 ($\Delta$ Top-1)} & \textbf{Pruned Top-5 ($\Delta$ Top-5)} & \textbf{Pruned FLOPs} \\
    \hline
    \multirow{7}{*}{MobileNet-V2} & Uniform\cite{b30} & 71.80\% & 91.00\% & 69.80\%(-2.00\%) & 89.60\%(-1.40\%) & 30.0\% \\
    \cline{2-7}
     & AMC\cite{b45} & 71.80\% & - & 70.80\%(-1.00\%) & - & 30.0\% \\
    \cline{2-7}
     & CC\cite{b46} & 71.88\% & - & 70.91\%(-0.97\%) & - & 28.3\% \\
    \cline{2-7}
     & MetaPruning\cite{b47} & 72.0\% & - & 71.20\%(-0.80\%) & - & 30.7\% \\
    \cline{2-7}
     & Random Pruning\cite{b48} & 71.88\% & - &70.87\%(-1.01\%) & - & 29.1\% \\
    \cline{2-7}
     & ATO\cite{b25} & 71.88\% & 90.29\% & 72.02\%(+0.14\%) & 90.19\%(-0.10\%) & 30.1\% \\
    \cline{2-7}
     & \textbf{CAMP-HiVe(Ours)} &\textbf{74.45\% }& \textbf{93.65\%} & \textbf{75.0\%(+0.55\%)} & \textbf{95.75\%(+0.1}\%) & \textbf{30.0\%} \\
    \hline
\end{tabular}
}
\vspace{-1mm}
\caption{Detailed comparison of accuracy and FLOPs reduction for different architectures on ImageNet dataset}
\label{imagenetDataResult}
\vspace{-6mm}
\end{table*}

\vspace{.5mm}
\noindent
\textbf{Performance on CIFAR-10 Dataset}\\
For evaluating the performance the selected models for CIFAR-10 are MobileNetV2, ResNet-56, ResNet-18.\ref{Cifat10_100_DataResult} represents the SOTA results along with our algorithm (i.e., Cyclic pair merge prune).

\textbf{ResNet-18:} 
In ResNet-18, we have achieved the best performance while reducing the most flops. We have achieved $\Delta$-Acc +0.43\%  in top-1 accuracy where flops reduction was 80\%. We compare our results with other SOTA results (table \ref{Cifat10_100_DataResult}) and show our approach outperforming them. 

\textbf{ResNet-56:} 
The results for ResNet-56 on the CIFAR-10 dataset demonstrate the effectiveness of the ``Cyclic paired merge pruning" method as well,
It provides a high level of accuracy after pruning, and a pruned model accuracy of 93.76\%. Additionally, it improves the baseline accuracy by +0.27\%, making it one of the best-performing methods in the comparison set. Other SOTA techniques, e.g., ATO and GNN-RL, DMC has also shown gain in accuracy, but our method achieved the most FLOPS reduction (60\%) along with the gain in  accuracy as well.

\textbf{MobileNet-V2:} 
In MobileNet-V2, our algorithm performs very well in terms of $\Delta$-Acc. It reduces almost the same percentage of FLOPS as \cite{b25}, but provides better performance in $\Delta$-Acc (+0.93\%).The detailed experimentation is performed with CIFAR-10 dataset to analyze the accuracy of the various pruning methods with respect the number of FLOPs. The cyclic pair merge pruning not only outperformed all the naive methods such as magnitude pruning, random merging but also the base model by around 2.5\%. We have compared our results with Magnitude pruning \cite{b26}, Hessian-based magnitude pruning where hessian vector has been used to identify the importance of the weight then Magnitude pruning was applied. We also have compared with Random Pair Merge Pruning where pairing between significant and less significant was done randomly as like random pruning \cite{b27} but with paring and merging then pruning. We have implemented all these methods and generated the results (see figure \ref{fig:AccuracyvsFlops} in section 5). Our approach has been stood out high and outperformed other approaches.


\noindent
\textbf{Performance on CIFAR-100 Dataset: } 
We have utilized ResNet-18 model to conduct the experiments on CIFAR-100 
as shown in Table\ref{Cifat10_100_DataResult}. Our proposed method has performed well with higher FLOPS reduction (50\%) and achieved better performance than the other SOTA pruning approaches in terms of accuracy and flop reduction.

\vspace{1mm}
\noindent
\textbf{Performance on ImageNet Dataset:}
Additionally, we also analyzed the performance of cyclic paired merge pruning with different existing approaches on ImageNet dataset for MobileNet-V2 model as shown in Table~\ref{imagenetDataResult}. The cyclic paired merge pruning achieved substantial improvement in Top-1, Top-5 accuracy for both the base and pruned models. In 30.0\% of FLOPS reduction, our approach has gain in both Top-1 ($\Delta$ +0.55\%) and Top-5 ($\Delta$ +0.1\%) accuracy which outperformed the other approaches.

\vspace{-2mm}
\subsection{\textbf{Feature Map Computation and Statistics}}
\vspace{-1mm}
We analyze the effect of unstructured weight pruning on feature retention across different layers of the MobileNetV2 and Resnet-56 models. Our approach evaluates how pruning affects the activation values of feature maps at different depths, providing an empirical understanding of information preservation across various pruning ratios (30\%, 50\%, 70\%, and 80\%).

\vspace{-1mm}
To quantitatively assess feature retention, we measure the minimum, maximum, and mean activation values of feature maps before and after pruning. The difference between the pruned and base models is captured using the Mean Absolute Deviation (MAD) metric, which quantifies the absolute difference in activation values across all spatial locations and channels. 

\begin{figure}[!t]
    \centering
    \vspace{-3mm}
    \includegraphics[scale=.30]{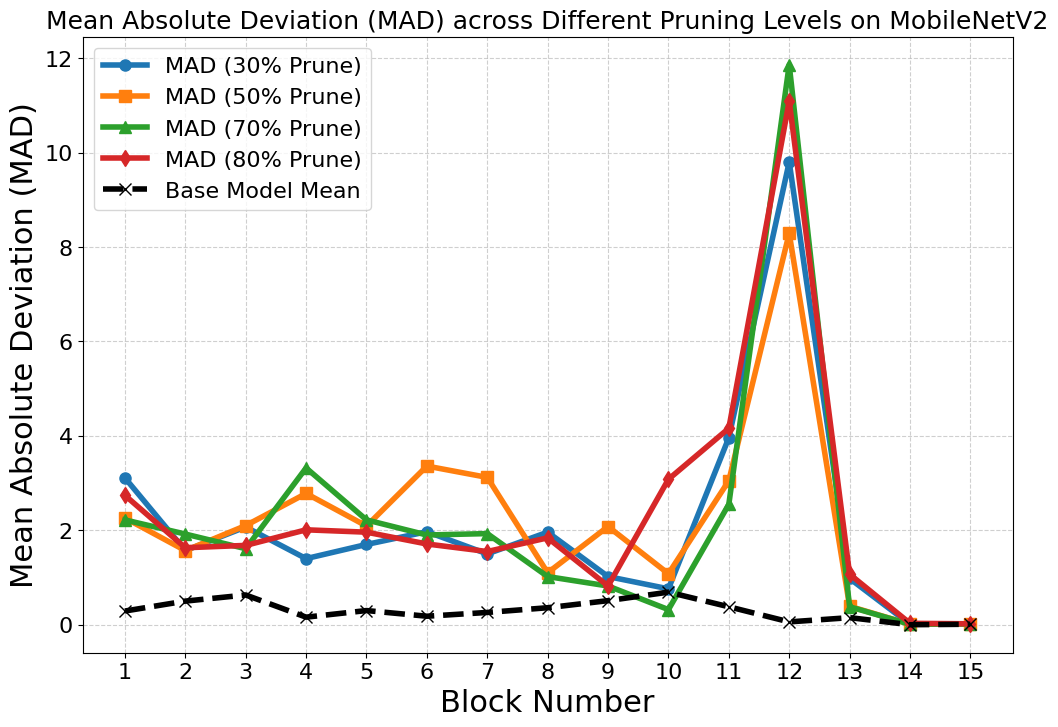}
    \vspace{-3mm}
    \caption{MAD across different pruning levels on MobileNetV2}
\label{fig:mobilenet_MAD_curve}
\vspace{-3mm}
\end{figure}

\vspace{-1mm}
In MobileNetV2 (see figure \ref{fig:mobilenet_MAD_curve}), early layers (Blocks 1-5) are responsible for extracting low-level spatial features such as edges and textures. Preserving weight connectivity in these layers is crucial to prevent cascading errors. Our analysis shows that for these layers, 30\% pruning introduces minimal activation deviation (MAD \textless{} 1.5), ensuring feature retention, while 50\% pruning eliminates redundancy effectively, retaining 85\%-90\% of the base model’s activations. With 70\%-80\% pruning, the network adapts by redistributing important weights, maintaining core feature representations.
In mid layers (Blocks 6-10), which encode structural relationships and object compositions, pruning optimally redistributes weight removal, ensuring stable activation distributions. Here, 
at 50\% pruning, we achieve an optimal balance between efficiency and feature integrity, maintaining MAD \textless{} 2.0, 
Notably, 
70\%-80\% pruning retains essential feature representations,
through efficient weight redistribution.
For deep layers (Blocks 11-16), which store high-level semantic and class-discriminative information, pruning strategies must ensure that weight sparsification does not reduce feature separability. In these layers, while 30\%-50\% pruning preserves critical activations and stable classification performance, 70\%-80\% pruning leverages sparsity-driven generalization, improving inference speed while retaining key class-discriminative features. 
\begin{figure}[!t]
    \centering
   \vspace{-2mm} 
   \includegraphics[width=1.05\linewidth]{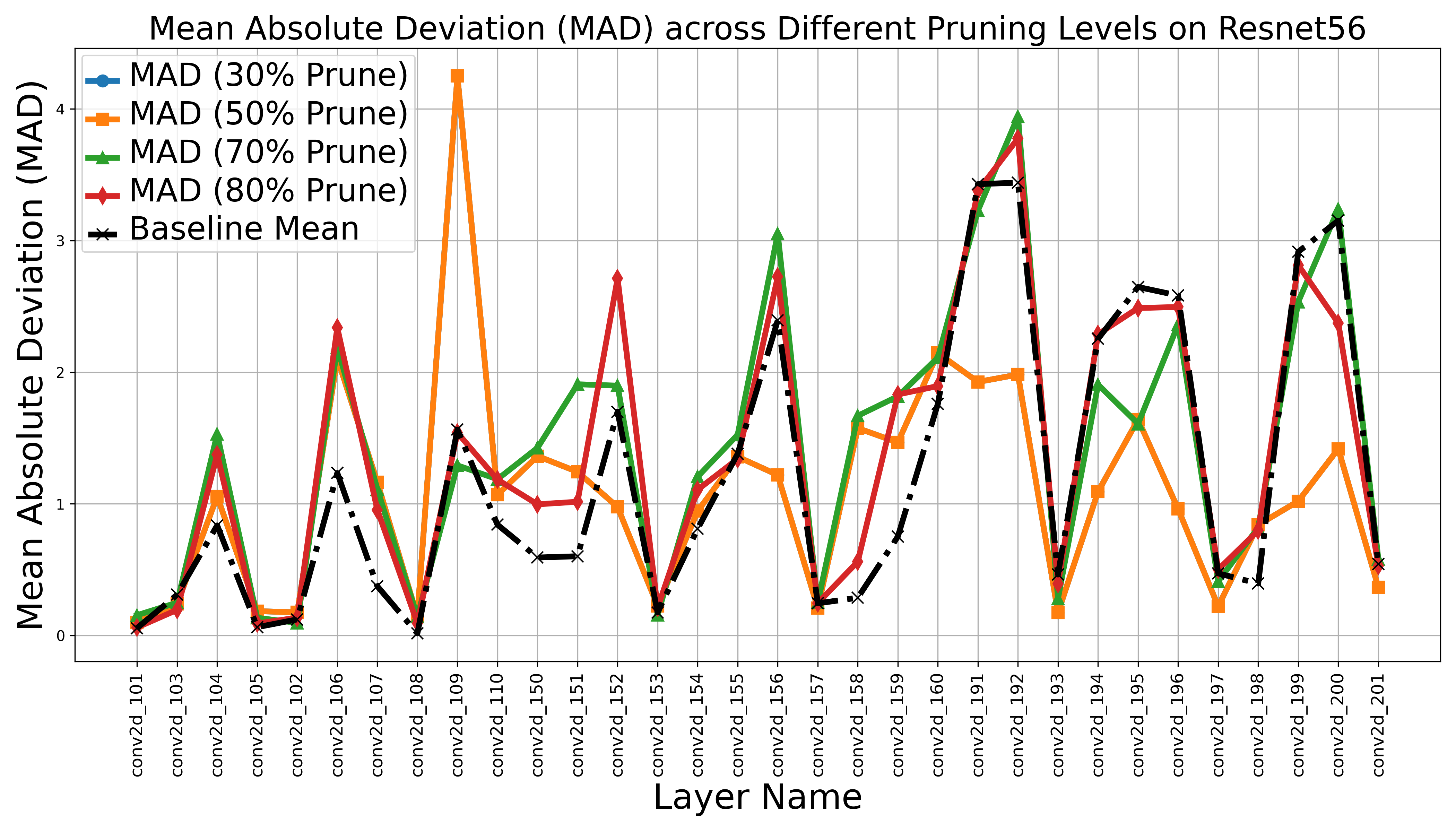}
    \vspace{-6mm}
    \caption{MAD across different pruning levels on Resnet-56}
\label{fig:resnet56_map_curve}
\end{figure}

\begin{figure*}[!t]
    \centering
    \includegraphics[width=1\linewidth]{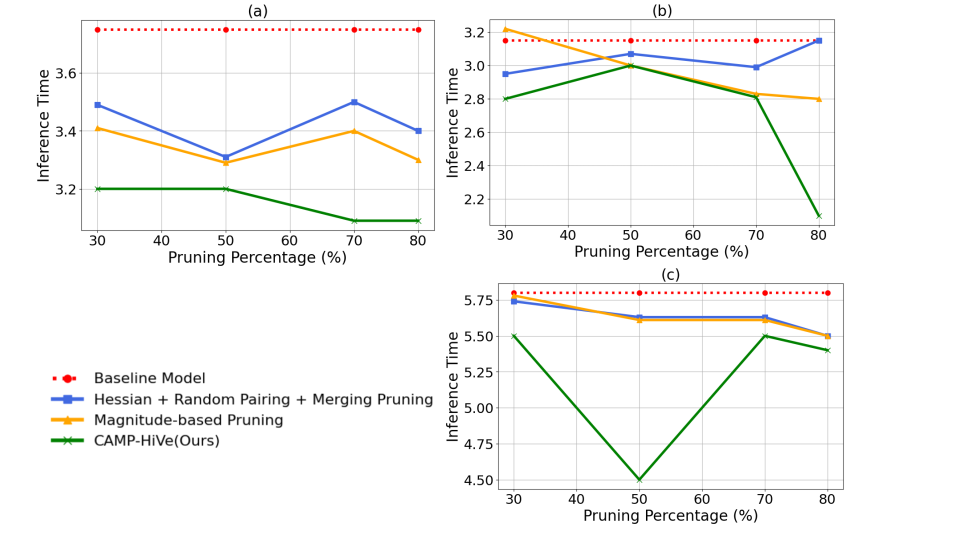}
    \vspace{-2mm}
    \caption{Inference vs pruning percentage comparison across different stages of pruned models on MobileNet-V2 for various edge devices( (a) Nvidia Jetson AGX-Orin 16GB, (b) Nvidia Jetson AGX-Orin-32GB, (c) Nvidia Jetson Orin-Nano-4GB).}
    \label{fig:combined_figure}
    \vspace{-4mm}
\end{figure*}

In ResNet-56 (see figure \ref{fig:resnet56_map_curve}), in the early layers (conv2d\_101 - conv2d\_110), responsible for edge detection and texture encoding, 30-50\% pruning preserves essential features with minimal distortion (e.g., conv2d\_106 MAD = 2.1066), while at 70\% pruning, the model begins adapting by selectively suppressing low-importance details, leading to a more compact representation of fundamental textures (conv2d\_103 mean: 0.5350 → 0.4267). At 80\% pruning, the network prioritizes the most critical edges and patterns, resulting in a focused yet lean feature set (conv2d\_109 MAD: 1.2923 → 1.5401). In the middle layers (conv2d\_150 - conv2d\_160), which refine and transform features, 30-50\% pruning maintains stable activations (conv2d\_152 MAD = 0.9762), while at 70\% pruning, the network dynamically adjusts, enhancing its ability to extract key patterns efficiently (conv2d\_156 MAD: 1.2193 → 3.0512). At 80\% pruning, the model exhibits a sharper focus on core representations, streamlining its feature encoding process (conv2d\_152 MAD: 1.8976 → 2.7120). In the late layers (conv2d\_191 - conv2d\_201), crucial for class-specific feature extraction, 30-50\% pruning retains classification stability (conv2d\_191 MAD = 1.9259). At 70\% pruning, the model refines its class separability by emphasizing the most distinctive features, leading to a more refined decision-making process (conv2d\_192 MAD: 1.9820 → 3.9429). By 80\% pruning, the network efficiently consolidates its final feature maps, maintaining strong classification performance while reducing redundancy (conv2d\_200 MAD: 1.4180 → 3.2399 → 2.3726). Overall, 30-50\% pruning maintains feature extraction and classification, while 70-80\% pruning enables the model to focus on the most critical patterns, promoting efficiency and compact representation with selective feature retention.


\vspace{.5mm}
\subsection{\textbf{Experiments on Edge Devices}}
\vspace{-1mm}
To test our proposed solution on real-world resource-constrained embedded systems, we test with the CIFAR10 dataset and the individual pruned models are loaded on Nvidia Jetson AGX Orin for inference. 
Since Tensorflow allocates GPU memory and does not immediately deallocate memory when it is no longer being used, we run the processes one at a time for evaluation.
Each process begins after a logging thread is started on the main process, gathering constant data from JTop (jetson-stats). The new process sends back timestamp data through a pipe at the beginning and end of model inference, as well as an idle period. Once the test on an individual model ends in this process, it is killed and releases all Tensorflow memory. The main process ends the log, saves it, and starts another log and process for the next model.

Each model was tested on the same dataset for three iterations, which allows us to identify outliers in the data. In these tests, individual iterations have their own log files, and are run in separate processes. Post-processing scripts combine these iterations for plotting data.
Models were tested on the following device configurations mentioned in Table \ref{tab:orin_devices}, however, due to limited space we report results for three of them here.

\begin{table}[!t]
\centering
\renewcommand{\arraystretch}{1.5} 
\resizebox{\columnwidth}{!}{
\begin{tabular}{|l|c|c|c|}
\hline
\Large\textbf{Device} & \Large\textbf{GPU Cores} & \Large\textbf{Max GPU Frequency} & \Large\textbf{Shared Memory} \\ \hline
\large AGX Orin & \large 1792 & \large 930 MHz & \large 32 GB \\ \hline
\large Orin NX & \large 1024 & \large 918 MHz & \large 16 GB \\ \hline
\large Orin Nano & \large 512 & \large 625 MHz & \large 4 GB \\ \hline
\large AGX Orin Devkit (old version) & \large 2048 & \large 1.3 GHz & \large 32 GB \\ \hline
\large Orin NX & \large 1024 & \large 765 MHz & \large 8 GB \\ \hline
\large Orin Nano & \large 1024 & \large 625 MHz & \large 8 GB \\ \hline
\end{tabular}
}
\vspace{-2mm}
\caption{Specifications of Various Orin Devices}
\label{tab:orin_devices} 
\end{table}

\begin{figure*}[!t]
    \centering
    \includegraphics[width=1\linewidth]{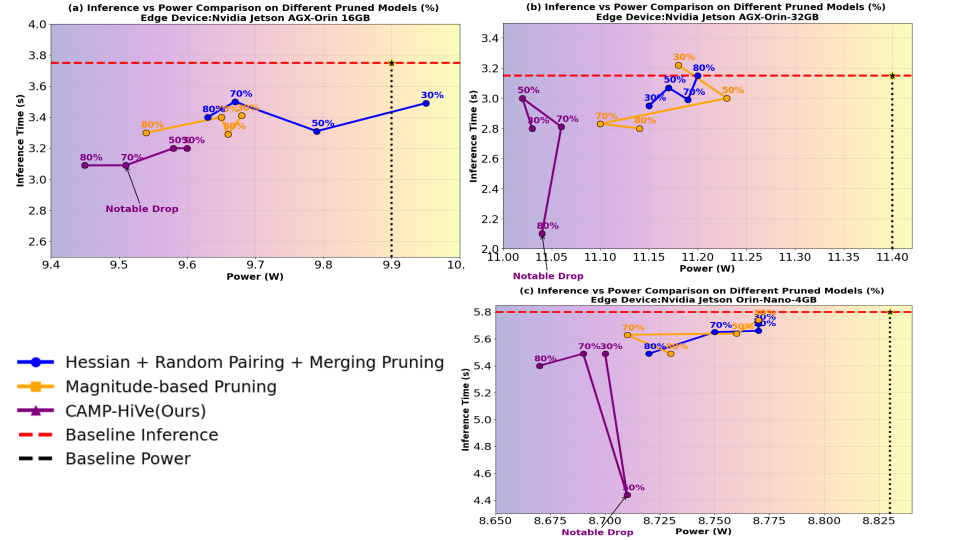}
    \vspace{-1mm}
    \caption{Power vs Inference comparison across different stages of pruned models on MobileNet-V2 for various edge devices: (a) Nvidia Jetson AGX-Orin 16GB, (b) Nvidia Jetson AGX-Orin-32GB, (c) Nvidia Jetson Orin-Nano-4GB.}
    \label{fig:combined_figure_powervsinference}
    \vspace{-5mm}
\end{figure*}

Figure~\ref{fig:combined_figure}, shows the inference latency vs different versions of pruned models with different degree of compression. MobileNet-V2 was used for this experiment across the different edge devices. In the two most resource-constrained devices, e.g., (b) and (c), our approach  always performed better than the others. In device (a) which as higher resources, at the lower and higher pruned regime our approach performed better than all others, whereas in mid-region it performed close to the "Magnitude pruning". 

In figure \ref{fig:combined_figure_powervsinference}, we have show comparison between power consumption and inference latency for the devices. The points towards the bottom left corner of the plot is better and upper-right corner shows worse performance. For any given compression percentage, our approach has shown better results than others, i.e., lower inference latency and/or lower power consumption. Especially on the device (a) Nvidia Jetson Orin-Nano 4GB, we have observed an impressive performance from our approach. 

\vspace{-3mm}
\section{\textbf{Ablation Study}}
\vspace{-2mm}
To thoroughly assess the impact of Cyclic Pair Merge Pruning (CAMP-HiVe), we conducted a series of ablation experiments that evaluate the effect of different pruning strategies, the importance of Hessian-based selection, weight redistribution techniques, performance across different sparsity levels, and computational efficiency on edge devices. 

\begin{figure}[!t]
    \centering
    \vspace{-2mm}
    \includegraphics[scale =0.21]{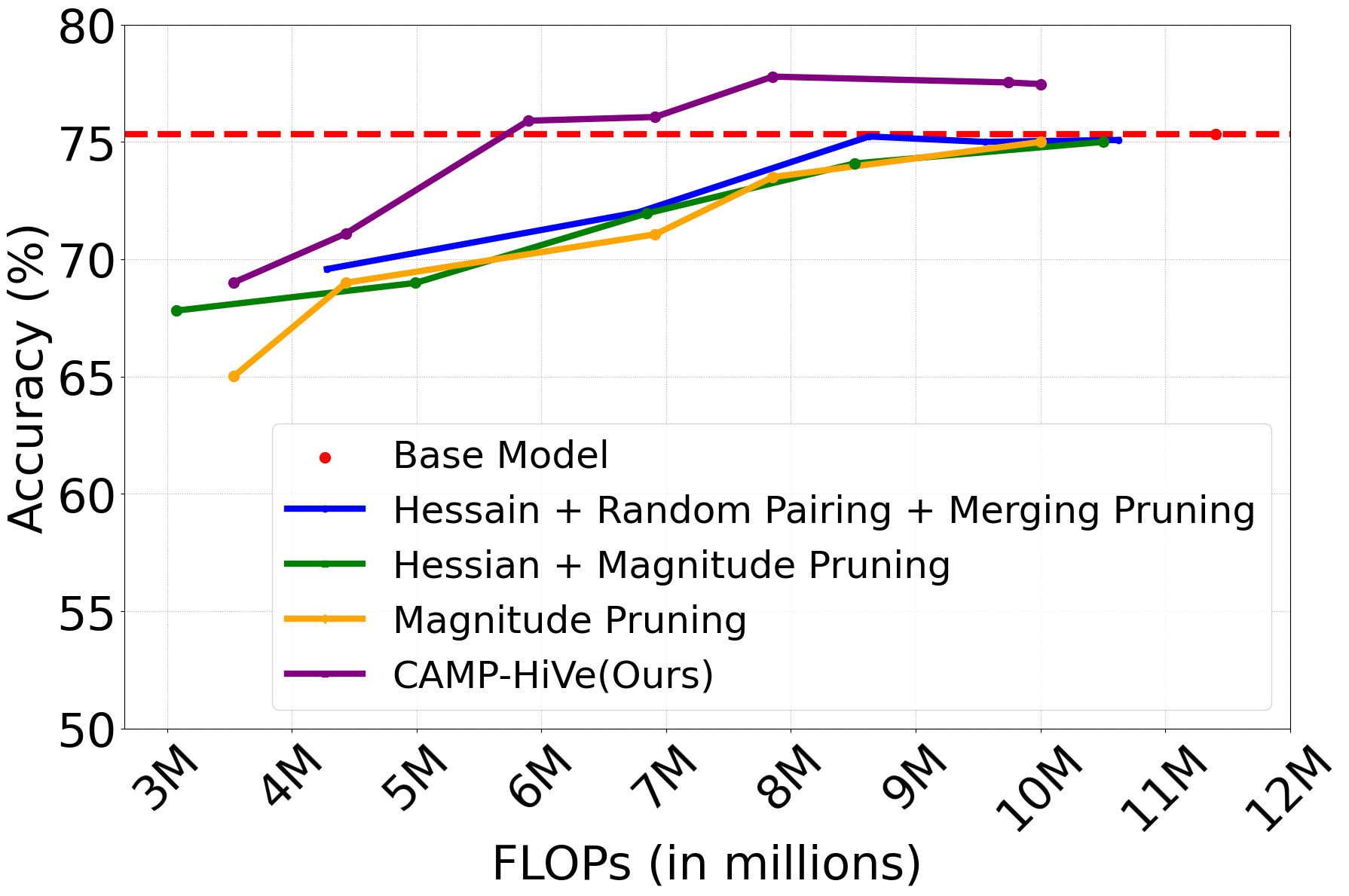}
    \vspace{-0.5mm}
    \vspace{-2mm}
    \caption{Accuracy vs FLOPS comparison with different kinds of models on MobileNet-V2 with CIFAR-10 dataset.}
\label{fig:AccuracyvsFlops}
\vspace{0.5mm}
\end{figure}

To validate the effectiveness of this approach, we compare CAMP-HiVe against - Hessian + Random Pairing + Merging Pruning (HRP), Hessian + Magnitude Pruning (HMP), Traditional Magnitude Pruning (MP), Baseline Model (MobileNet-V2), figure \ref{fig:AccuracyvsFlops}. In our experiments, CAMP-HiVe retains 76.06\% accuracy at 50\% pruning, outperforming HMP, and MP, which achieve 71.95\% and 71.06\%, respectively.CAMP-HiVe retains significantly higher accuracy compared to other methods, demonstrating that cyclic pairing enhances weight retention efficiency and prevents excessive loss of critical parameters. By leveraging Hessian-vector products for importance estimation, CAMP-HiVe ensures that pruning decisions are influenced by second-order gradients, leading to smarter weight selection.


CAMP-HiVe outperforms Magnitude Pruning by 5\% at the same pruning rate, demonstrating the impact of Hessian-based importance estimation in preserving critical connections. At 70\% pruning, it achieves 3.3\% higher accuracy than HMP, validating that weight redistribution helps retain structural integrity and enhances robustness. Unlike abrupt sparsity-inducing methods, CAMP-HiVe’s merging strategy promotes stable parameter utilization. Its robustness is further confirmed by consistent accuracy retention across various pruning ratios (figure \ref{fig:AccuracyvsFlops}).

\begin{table}[!b]
\vspace{3mm}
\centering
\renewcommand{\arraystretch}{2} 
\resizebox{\columnwidth}{!}{
\begin{tabular}{|l|c|c|c|}
\hline
\Large\textbf{Pruning \%} & \Large\textbf{Orin 16GB} & \Large\textbf{Orin 32GB} & \Large\textbf{Orin Nano 4GB} \\ \hline
\large Baseline Model & \large 3.75 ms & \large 3.15 ms & \large 5.8 ms \\ \hline
\large CAMP-HiVe (Ours) - 50\% Pruning	 & \large 3.2 ms (-14.6\%)	 & \large 3.0 ms (-4.7\%)& \large 4.44 ms (-23.4\%) \\ \hline
\large CAMP-HiVe (Ours) - 70\% Pruning	 & \large 3.09 ms (-17.6\%) & \large 2.81 ms (-10.8\%) & \large 5.49 ms (-5.3\%)
 \\ \hline

\end{tabular}
}
\caption{Inference differences of different models on Edge devices.}
\label{tab:interence_vspruning} 
\end{table}

To evaluate real-world applicability, we assess inference time and power consumption on three NVIDIA Jetson edge devices in table \ref{tab:interence_vspruning} shows that CAMP-HiVe’s ability to reduce inference latency while maintaining high accuracy. The largest speedup is observed on the Orin Nano 4GB device, where inference time is reduced by 23.4\% at 50\% pruning, showcasing CAMP-HiVe’s efficiency in resource-constrained environments, also table \ref{tab:interence_vspower} demonstrate that CAMP-HiVe not only improves inference speed but also reduces power consumption across all tested devices. The Orin 16GB model sees up to 3.9\% lower power usage, while the Orin 32GB and Orin Nano 4GB achieve 3.0\% and 1.6\% reductions, respectively. This confirms CAMP-HiVe’s efficiency in balancing power and computational performance.

\begin{table}[!t]
\vspace{2mm}
\centering
\renewcommand{\arraystretch}{2} 
\resizebox{\columnwidth}{!}{
\begin{tabular}{|l|c|c|c|}
\hline
\Large\textbf{Pruning \%} & \Large\textbf{Orin 16GB (W)} & \Large\textbf{Orin 32GB (W)} & \Large\textbf{Orin Nano 4GB (W)} \\ \hline
\large Baseline Model & \large 9.9 W & \large 11.40 W & \large 8.83 W \\ \hline
\large CAMP-HiVe (Ours) - 50\% Pruning	 & \large 9.58 W (-3.2\%)	 & \large 11.02 W (-3.3\%)& \large 8.71 W (-1.4\%) \\ \hline
\large CAMP-HiVe (Ours) - 70\% Pruning	 & \large 9.51 W (-3.9\%)	 & \large 11.06 W (-3.0\%) & \large 8.69 W (-1.6\%)
 \\ \hline

\end{tabular}
}
\caption{Power consumption of different models on Edge devices.}
\label{tab:interence_vspower} 
\end{table}

\vspace{-1mm}
\section{\textbf{Conclusion}}
\vspace{-1mm}
In this work, we have introduced a novel neural network pruning approach namely Cyclic Paired Merge Pruning which can effectively reduce the neural network complexity while without degrading or minimally degrading the performance. Our solution utilizes Hessian-vector products and a novel cyclic pairing strategy to efficiently merge less significant weights into more crucial ones which preserves the model integrity. The experimental findings showed that our technique outperforms the state-of-the-art algorithms on standard datasets CIFAR-10, CIFAR-100, and ImageNet, achieving significant reductions in computational demands without compromising accuracy. Our technique also improves the edge device inference speed and power efficiency, making it ideal for resource-constrained systems and applications.

\section{ACKNOWLEDGMENTS}
This work is partially supported by the US National Science Foundation (EPSCoR \#1849213).

\vspace{12pt}
\color{red}


\begin{thebibliography}{00}
\bibitem{b1} G. Eason, B. Noble, and I. N. Sneddon, ``On certain integrals of Lipschitz-Hankel type involving products of Bessel functions,'' Phil. Trans. Roy. Soc. London, vol. A247, pp. 529--551, April 1955.
\bibitem{b2} Y. LeCun, Y. Bengio, and G. Hinton, “Deep learning,” Nature, vol. 521, no. 7553, pp. 436–444, 2015.
\bibitem{b3} A. Voulodimos, N. Doulamis, A. Doulamis, and E. Protopapadakis, “Deep learning for computer vision: A brief review,” Comput. Intell. Neurosci., vol. 2018, 2018.
\bibitem{b4} R. Szeliski, Computer Vision: Algorithms and Applications. Cham, Switzerland: Springer Nature, 2022.
\bibitem{b5} T. Wolf et al., “Transformers: State-of-the-art natural language processing,” in Proc. 2020 Conf. Empirical Methods in Natural Language Processing: System Demonstrations, 2020, pp. 38–45.
\bibitem{b6}  K. R. Chowdhary, “Natural language processing,” in Fundamentals of Artificial Intelligence. Springer, 2020, pp. 603–649.
\bibitem{b7} S. Rajbhandari, J. Rasley, O. Ruwase, and Y. He, “Zero: Memory optimizations toward training trillion parameter models,” in Proc. SC20: Int. Conf. High Performance Computing, Networking, Storage and Analysis, 2020, pp. 1–16.
\bibitem{b8} Y. LeCun, J. Denker, and S. Solla, “Optimal brain damage,” in Adv. Neural Inf. Process. Syst., vol. 2, 1989.
\bibitem{b9}M. C. Mozer and P. Smolensky, “Skeletonization: A technique for trimming the fat from a network via relevance assessment,” in Adv. Neural Inf. Process. Syst., vol. 1, 1988.
\bibitem{b10} J. Martens and R. Grosse, “Optimizing neural networks with Kronecker-factored approximate curvature,” in Proc. Int. Conf. Machine Learning (ICML), 2015, pp. 2408–2417

\bibitem{b11} J. Frankle and M. Carbin, “The Lottery Ticket Hypothesis: Finding Sparse, Trainable Neural Networks,” in Proc. Int. Conf. Learn. Representations (ICLR), 2019.
\bibitem{b12} A. S. Morcos, H. Yu, M. Paganini, and Y. Tian, “One Ticket to Win Them All: Generalizing Lottery Ticket Initializations Across Datasets and Optimizers,” in Adv. Neural Inf. Process. Syst. (NeurIPS), 2019.
\bibitem{b13} A. Renda, J. Frankle, and M. Carbin, “Comparing Rewinding and Fine-Tuning in Neural Network Pruning,” in Proc. Int. Conf. Learn. Representations (ICLR), 2020.
\bibitem{b16} Y. He, G. Kang, X. Dong, Y. Fu, and Y. Yang, “Soft Filter Pruning for Accelerating Deep Convolutional Neural Networks,” in Proc. Int. Joint Conf. Artif. Intell. (IJCAI), 2019.
\bibitem{b17} M. Lin, R. Ji, Y. Zhang, B. Zhang, Y. Wu, and Y. Tian, “HRank: Filter Pruning Using High-Rank Feature Map,” in Proc. IEEE/CVF Conf. Comput. Vis. Pattern Recognit. (CVPR), 2020.
\bibitem{b18} V. Sanh, L. Debut, J. Chaumond, and T. Wolf, “DistilBERT, a Distilled Version of BERT: Smaller, Faster, Cheaper and Lighter,” in Adv. Neural Inf. Process. Syst. (NeurIPS), 2019.
\bibitem{b19} R. Tang, Y. Lu, L. Liu, L. Mou, O. Vechtomova, and J. Lin, “TinyBERT: Distilling BERT for Natural Language Understanding,” in Findings Assoc. Comput. Linguistics: EMNLP, 2020.
\bibitem{b20} C. Wang, R. Grosse, S. Fidler, and G. Zhang, "EigenDamage: Structured pruning in the Kronecker-factored eigenbasis," in Proc. 36th Int. Conf. Mach. Learn. (ICML), vol. 97, 2019, pp. 6566–6575.
\bibitem{b21} X. Dong, S. Chen, and S. J. Pan, “Scalable Hessian Approximation for Neural Network Pruning,” in Proc. ICLR, 2022.
\bibitem{b22} B. A. Pearlmutter, “Fast Exact Multiplication by the Hessian,” Neural Computation, vol. 6, no. 1, pp. 147–160, 1994.
\bibitem{b23} J. Martens, “Deep Learning via Hessian-Free Optimization,” in Proc. 27th Int. Conf. Mach. Learn. (ICML), 2010, pp. 735–742. [Online]. Available: http://www.icml2010.org/papers/458.pdf
\bibitem{b24} I. Goodfellow, Y. Bengio, and A. Courville, Deep Learning. Cambridge, MA: MIT Press, 2016. [Online]. Available: https://www.deeplearningbook.org/

\bibitem{b25} X. Wu, S. Gao, Z. Zhang, Z. Li, R. Bao, Y. Zhang, X. Wang, and H. Huang, “Auto-Train-Once: Controller Network Guided Automatic Network Pruning from Scratch,” arXiv preprint arXiv:2403.14729, 2024. 
\bibitem{b26} S. Han, J. Pool, J. Tran, and W. Dally, “Learning Both Weights and Connections for Efficient Neural Network,” in Adv. Neural Inf. Process. Syst. (NeurIPS), vol. 28, 2015.
\bibitem{b27} J. Frankle, G. K. Dziugaite, D. M. Roy, and M. Carbin, “Linear Mode Connectivity and the Lottery Ticket Hypothesis,” in Proc. Int. Conf. Mach. Learn. (ICML), 2021, pp. 3259–3270.
\bibitem{b28} T. Chen, L. Liang, T. Ding, Z. Zhu, and I. Zharkov, “Otov2: Automatic, Generic, User-Friendly,” arXiv preprint arXiv:2303.06862, 2023. 

\bibitem{b30} Z. Zhuang, M. Tan, B. Zhuang, J. Liu, Y. Guo, Q. Wu, J. Huang, and J. Zhu, “Discrimination-Aware Channel Pruning for Deep Neural Networks,” in Adv. Neural Inf. Process. Syst., 2018, pp. 875–886.
\bibitem{b31} M. Kang and B. Han, “Operation-Aware Soft Channel Pruning Using Differentiable Masks,” in Proc. Int. Conf. Mach. Learn. (ICML), 2020.
\bibitem{b32}Y. He, P. Liu, Z. Wang, Z. Hu, and Y. Yang, “Filter Pruning via Geometric Median for Deep Convolutional Neural Networks Acceleration,” in Proc. IEEE Conf. Comput. Vis. Pattern Recognit. (CVPR), 2019.
\bibitem{b33} Y. He, G. Kang, X. Dong, Y. Fu, and Y. Yang, “Soft Filter Pruning for Accelerating Deep Convolutional Neural Networks,” in Proc. Int. Joint Conf. Artif. Intell. (IJCAI), 2018, pp. 2234–2240.
\bibitem{b34} Y. He, Y. Ding, P. Liu, L. Zhu, H. Zhang, and Y. Yang, “Learning Filter Pruning Criteria for Deep Convolutional Neural Networks Acceleration,” in Proc. IEEE/CVF Conf. Comput. Vis. Pattern Recognit. (CVPR), 2020.
\bibitem{b36} S. Gao, F. Huang, J. Pei, and H. Huang, “Discrete Model Compression with Resource Constraint for Deep Neural Networks,” in Proc. IEEE/CVF Conf. Comput. Vis. Pattern Recognit. (CVPR), 2020.

\bibitem{b37} S. Yu, A. Mazaheri, and A. Jannesari, “Topology-Aware Network Pruning Using Multi-Stage Graph Embedding and Reinforcement Learning,” in Proc. Int. Conf. Mach. Learn. (ICML), 2022.


\bibitem{b40} Y. Tang, Y. Wang, Y. Xu, D. Tao, C. Xu, C. Xu, and C. Xu, “SCOP: Scientific Control for Reliable Neural Network Pruning,” in Adv. Neural Inf. Process. Syst., vol. 33, 2020.

\bibitem{b45} Y. He, J. Lin, Z. Liu, H. Wang, L.-J. Li, and S. Han, “AMC: AutoML for Model Compression and Acceleration on Mobile Devices,” in Proc. Eur. Conf. Comput. Vis. (ECCV), 2018.
\bibitem{b46} Y. Li, S. Lin, J. Liu, Q. Ye, M. Wang, F. Chao, F. Yang, J. Ma, Q. Tian, and R. Ji, “Towards Compact CNNs via Collaborative Compression,” in Proc. IEEE/CVF Conf. Comput. Vis. Pattern Recognit. (CVPR), 2021.
\bibitem{b47} Z. Liu, H. Mu, X. Zhang, Z. Guo, X. Yang, K.-T. Cheng, and J. Sun, “Metapruning: Meta Learning for Automatic Neural Network Channel Pruning,” in Proc. IEEE Int. Conf. Comput. Vis. (ICCV), 2019.
\bibitem{b48} Y. Li, K. Adamczewski, W. Li, S. Gu, R. Timofte, and L. Van Gool, “Revisiting Random Channel Pruning for Neural Network Compression,” in Proc. IEEE/CVF Conf. Comput. Vis. Pattern Recognit. (CVPR), 2022.
\bibitem{b49} P. Molchanov, S. Tyree, T. Karras, T. Aila, and J. Kautz, “Pruning convolutional neural networks for resource-efficient inference,” in Proc. Int. Conf. Learn. Representations (ICLR), Toulon, France, Apr. 2017.



\end{thebibliography}
\end{document}